\documentclass[11pt,a4paper]{article}
\usepackage{ranlp2025}
\usepackage{amsmath}
\usepackage{times}
\usepackage{latexsym}

\usepackage{amsmath}
\usepackage{amsfonts}
\usepackage{natbib}
\usepackage{hyperref}
\usepackage{booktabs}
\usepackage{adjustbox}
\usepackage{microtype}
\usepackage{graphicx}
\aclfinalcopy

\usepackage[most]{tcolorbox}
\tcbset{
  colback=gray!5,
  colframe=black!60,
  boxrule=0.5pt,
  arc=2mm,
  left=5pt,
  right=5pt,
  top=4pt,
  bottom=4pt,
  boxsep=4pt,
  enhanced,
}
\title{Steering Towards Fairness: Mitigating Political Bias in LLMs}

\author{
  Afrozah Nadeem, Mark Dras, Usman Naseem \\
  School of Computing, Macquarie University, Australia \\
  \texttt{afrozah.nadeem@students.mq.edu.au}, \\\texttt{\{mark.dras,usman.naseem\}@mq.edu.au}
}

\date{}

\begin{document}
\maketitle
\begin{abstract}
Recent advancements in large language models (LLMs) have enabled their widespread use across diverse real-world applications. However, concerns remain about their tendency to encode and reproduce ideological biases along political and economic dimensions. In this paper, we employ a framework for probing and mitigating such biases in decoder-based LLMs through analysis of internal model representations. Grounded in the Political Compass Test (PCT), this method uses contrastive pairs to extract and compare hidden layer activations from models like Mistral and DeepSeek. We introduce a comprehensive activation extraction pipeline capable of layer-wise analysis across multiple ideological axes, revealing meaningful disparities linked to political framing. Our results show that decoder LLMs systematically encode representational bias across layers, which can be leveraged for effective steering vector-based mitigation. This work provides new insights into how political bias is encoded in LLMs and offers a principled approach to debiasing beyond surface-level output interventions. 
\end{abstract}

\section{Introduction}

Large Language Models (LLMs) have become foundational tools across a wide spectrum of applications, yet their outputs frequently reflect political and ideological biases, particularly in contexts involving sensitive framing or policy-oriented discourse \cite{zheng_judging_2023, afzoon2025exbigbang}. This problem is particularly pressing in multilingual low-resource settings, where LLMs often produce uneven or culturally misaligned outputs across different languages, amplifying social or political asymmetries \cite{kumar_language_2023, maskey2025safeconstellations}.

\begin{figure}[!t]
\centering
\includegraphics[width=0.85\linewidth]{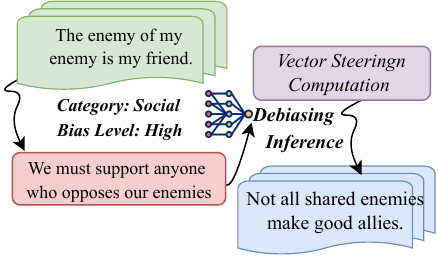}
\caption{Example of social bias mitigation in our framework. The input PCT statement (4) triggers a high-bias response aligned with tribal loyalty framing.}
\label{fig:intro_example}
\end{figure}

Emerging research reveals that a model’s ideological leanings are more influenced by input language than by intended sociocultural identity, raising serious concerns about fairness in multilingual settings \cite{helwe2025navigating}. For instance, the same political statement can elicit starkly different responses when phrased in Urdu versus Punjabi, even within the same model. As illustrated in Figure~\ref{fig:intro_example}, such biases can result in overconfident responses that reflect tribal or populist framings, potentially skewing downstream interpretations.

Prior studies have largely focused on evaluating LLM bias at the output level—either by quantifying stance across Political Compass Test (PCT) statements \cite{barkhordar_why_nodate} or by cataloging surface-level disparities across languages. However, these approaches stop short of proposing effective and reproducible mitigation strategies that operate within the internal representation space of decoder models \cite{ejaz_politics_2023}.

To address this gap, we investigate a modular activation-based mitigation framework that uses contrastive ideological prompts from the PCT to extract, analyze, and intervene on latent bias directions within decoder LLMs. At the core of the method is the use of Steering Vector Ensembles (SVE): layer-specific vector representations that capture ideological framing and allow for inference-time debiasing without fine-tuning \cite{siddique_shifting_2025}. Our contributions are as follows:

\begin{itemize}
\item We present a multilingual bias mitigation method on PCT using Steering Vector Ensembles derived from contrastive political prompts along social and economic axes.
\item Our pipeline supports scalable extraction and aggregation of hidden activations across decoder LLMs (e.g., Mistral, DeepSeek) in low-resource languages.
\item We demonstrate that ensemble-based interventions reduce bias while maintaining fluency and context relevance, offering a reproducible path toward fairer multilingual LLM behavior.
\end{itemize}
\begin{figure*}
    \centering
    \includegraphics[width=0.9\linewidth]{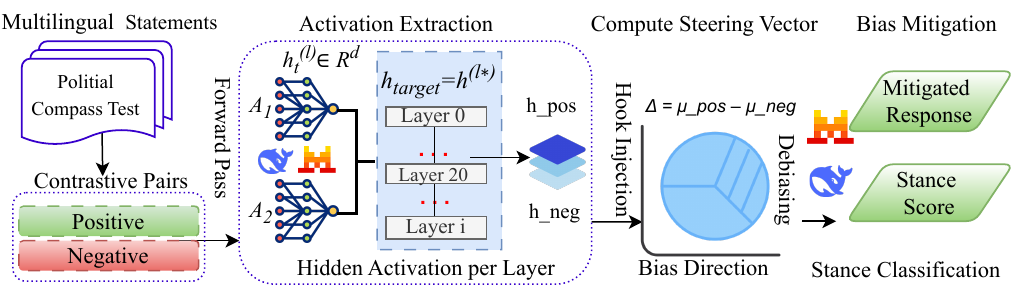}
    \caption{
        \textbf{Bias mitigation pipeline using steering vectors in transformer-based LLMs.}
        Multilingual statements from the Political Compass Test (PCT) are used to construct contrastive pairs representing opposing ideological stances (positive vs negative). Each pair is passed through a pretrained language model (e.g., DeepSeek-7B), and hidden states \( h^{(l)} \in \mathbb{R}^d \) are extracted from each transformer layer. A target layer \( l^* \) (e.g., 20) is selected, and its activations are mean-pooled to form \( h_{\text{pos}} \) and \( h_{\text{neg}} \). A bias direction vector is computed as \( v = \mu_{\text{pos}} - \mu_{\text{neg}} \), representing the difference between positive and negative class means. This vector is injected via a forward hook into the model's target layer. The modified model then generates a \textit{mitigated response}, which is evaluated using zero-shot stance classification to obtain a final \textit{stance score} \cite{thapa_which_nodate}.
    }
    \label{fig:bias_pipeline}
\end{figure*}

\section{Related Work}
\subsection{Bias Evaluation via the Political Compass Test (PCT)}
The Political Compass Test (PCT) has become a widely used diagnostic tool for probing the political leanings of LLMs \cite{helwe2025navigating}. Its structured two-axis framework—\textit{economic} (left–right) and \textit{social} (authoritarian–libertarian)—makes it particularly useful for assessing ideological alignment in model responses \cite{lee_neus_2022}.

Early studies \cite{liu_aligning_2024} leveraged the PCT for output-level bias evaluation, prompting models with ideologically framed statements and analyzing completions via stance classification or sentiment scoring. These studies uncovered consistent political leanings in popular LLMs, often skewing toward left-libertarian quadrants \cite{shen_large_2023}.

\paragraph{Multilingual Political Bias Studies.}
Recent research has highlighted that language plays a key role in shaping LLM bias. \citet{thapa_assessing_2023} translated the PCT into Nepali and found that smaller models exhibited economic-right bias, while larger ones leaned socially left. \citet{nadeem2025probing} extended this analysis to low-resource languages (Urdu and Punjabi), showing that models exhibited stronger authoritarian tendencies when generating in low-resource regional languages. Similarly, \citet{helwe2025navigating} evaluated 15 multilingual LLMs across 50 countries and found that both prompting language and persona assignment significantly influenced model stance—often more so than the nominal national identity \cite{feng_pretraining_nodate}.

These findings collectively underscore that political bias in LLMs is both pervasive and language-conditioned, and that multilingual evaluation is essential to uncovering such disparities. However, all these approaches remain post-hoc, focused solely on surface-level output, and do not probe the internal representation space where ideological bias is likely encoded.

\subsection{Steering Vectors and Ensemble Approaches for Mitigation}
Beyond evaluation, recent research has explored representation-level mitigation via steering vectors—directional vectors derived from hidden state differences between biased and neutral (or contrastive) inputs. Introduced in contexts like toxicity filtering and sentiment control \cite{sun_bertscore_2022}, steering vectors operate at the embedding or hidden state level, modifying a model’s response without retraining.

More recent work introduced Steering Vector Ensembles (SVE) \cite{siddique_shifting_2025}, which aggregate vectors across multiple demographic groups, model layers, or task settings. These ensembles offer improved robustness and generalizability. However, SVE studies have been narrow in scope, often focusing on: \textit{Encoder or encoder-decoder architectures like BERT or T5;}, \textit{Domain-specific settings, such as toxicity or fairness in QA;}, and \textit{English-only applications}, with little attention to ideological framing or multilingual dynamics. Thus, the potential of SVE for open-ended political discourse, particularly in decoder LLMs, remains largely unexplored.

Despite promising advancements, three core gaps remain in the literature:

\begin{itemize}
\item \textbf{Representation-level blind spots in decoder LLMs:} Most bias studies focus on outputs, leaving open questions about how and where ideological bias is encoded in decoder-only models like Mistral or DeepSeek \cite{rottger2024political}.
\item \textbf{Lack of systematic contrastive activation pipelines:} There is no open-source or standardized pipeline for extracting contrastive activations (e.g., liberal vs. authoritarian) across layers and prompts in decoder LLMs, particularly for multilingual bias detection.

\item \textbf{Underutilization of SVE in political contexts:} Steering Vector Ensembles have shown promise in fairness-related domains, but their application to ideological bias mitigation, particularly across languages and political axes, remains under-investigated \cite{chen_analyzing_2020}.
\end{itemize}

To address these limitations, our work introduces an activation-based bias mitigation pipeline tailored for decoder LLMs. Our core contributions include: A scalable, multilingual framework for layer-wise activation extraction using PCT-based contrastive pairs; A representation-level analysis method that identifies and aggregates ideological bias directions. And the first integration of Steering Vector Ensembles into decoder-based LLMs for mitigating political bias across both social and economic axes. By bridging output-based evaluation and internal mitigation strategies, we provide a new foundation for probing and correcting ideological bias in multilingual generative models. Although we focus on Political Compass Test (PCT) prompts, our pipeline is modular and could be extended to other domains such as healthcare or education. The ensemble design also improves robustness to prompt framing by aggregating across multiple paraphrases. Our implementation is publicly available to support reproducibility and further work \footnote{\url{https://github.com/Afx-Msh/SVE_Mitigation}}.

\section{Methodology}
\label{sec:methodology}
This section presents a framework for mitigating political bias in multilingual large language models (LLMs) using contrastive political pairs derived from the Political Compass Test (PCT). The pipeline integrates contrastive pair construction, activation-based analysis, and vector steering. We evaluate steering effectiveness using \textbf{Bias Score Reduction} ($\Delta$Bias) and response quality measures inspired by the work ~\cite{siddique_shifting_2025}.
\subsection{Framework Overview}
We introduce a modular pipeline for political debiasing of autoregressive large language models (LLMs) using contrastive prompting and steering vector interventions. Our approach consists of four main stages: \textit{constructing ideologically contrastive prompt pairs} based on translated Political Compass Test (PCT) statements, \textit{extracting hidden activations} from selected transformer layers, \textit{training layer-specific classifiers} to obtain directionally meaningful steering vectors, and \textit{injecting those vectors} during generation to modulate bias. 

We implement two steering strategies: \textit{Individual Steering Vectors (ISV)}, where a single vector is derived per layer using logistic regression, and \textit{Steering Vector Ensembles (SVE)}, where vectors from multiple layers are aggregated using response quality-weighted coefficients.

\subsection{Multilingual PCT Dataset Preparation}

We build on the multilingual PCT dataset proposed by \citet{nadeem2025probing}, which adapts the 62 standard Political Compass Test (PCT) statements into low-resource languages: Urdu and Punjabi \cite{smith_im_2022}. We extend this dataset to include English, resulting in six total languages spanning multiple language families \cite{mostefa_new_nodate}. 

Each translation was reviewed and verified by regional native speakers, achieving near-perfect inter-annotator agreement with a Fleiss' $\kappa = 0.99$.

The PCT statements cover both ideological axes:
\begin{itemize}
    \item \textbf{Economic axis}: left–right orientation (e.g., redistribution, market policies)
    \item \textbf{Social axis}: libertarian–authoritarian values (e.g., social freedoms, censorship)
\end{itemize}

Each statement was transformed into a pair of opposing ideological prompts through manual rewriting or structured agreement templates. To ensure semantic divergence and ideological contrast:
\begin{enumerate}
    \item We computed multilingual sentence embeddings using \texttt{sentence-transformers}.
    \item Contrastive pairs with cosine similarity below a threshold $\tau = 0.15$ were retained.
    \item We limited generation to a maximum of 30 pairs per category, with at most 500 comparisons, to avoid redundancy and infinite pairing loops.
\end{enumerate}

\subsection{Target Model and Layer Selection}

We selected \texttt{deepseek-llm-7b-chat}\footnote{\url{https://huggingface.co/deepseek-ai/deepseek-vl-7b-chat}}, Mistral model due to its strong multilingual capabilities and transparent architecture. We also evaluate \texttt{Mistral-7B-v0.1}\footnote{\url{https://huggingface.co/mistralai/Mistral-7B-v0.1}} to compare bias behavior across model families. We selected layers {8, 12, 16, 20, 24}, as layer-wise profiling indicated that mid-level layers encode the strongest ideological signals, whereas early layers capture lexical patterns and very late layers primarily influence fluency.

\subsubsection{Individual Steering Vectors (ISV)}

We compute a bias-aligned steering vector $\mathbf{v}_l$ for each selected transformer layer $l$ and each ideological axis. 

First, we extract hidden activations for positive (e.g., left-leaning) and negative (e.g., right-leaning) prompts. These are standardized using \texttt{StandardScaler}, and concatenated to form the input matrix $\mathbf{X} = [\mathbf{A}_{\text{pos}}; \mathbf{A}_{\text{neg}}]$. Corresponding binary labels are assigned as $\mathbf{y} = [1^{n_{\text{pos}}}; 0^{n_{\text{neg}}}]$.

Next, we train a logistic regression classifier with \texttt{max\_iter=1000} and \texttt{random\_state=42} to separate the two ideological classes. The resulting classifier weight vector $\boldsymbol{\theta}$ is normalized to unit length to obtain the steering vector $\mathbf{v}_l = \boldsymbol{\theta} / \| \boldsymbol{\theta} \|$.

Finally, to ensure directional consistency, we verify that the expected projection of positive activations exceeds that of negative activations, i.e., $\mathbb{E}[\mathbf{A}_{\text{pos}} \cdot \mathbf{v}_l] > \mathbb{E}[\mathbf{A}_{\text{neg}} \cdot \mathbf{v}_l]$.
\subsubsection{Quality Assessment}

Vector quality scores are computed as:

\begin{equation}
q_l = 0.6 \times \text{accuracy}_l + 0.4 \times \min\left( \frac{\text{separation}_l}{2},\ 1.0 \right)
\label{eq:quality_score}
\end{equation}

The term $\text{separation}_l$ captures the normalized effect size between projected activations of opposing ideological prompts and is computed as:

\begin{equation}
\text{separation}_l = \frac{|\mu_{\text{pos}} - \mu_{\text{neg}}|}{\text{pooled\_std}}
\label{eq:separation}
\end{equation}

where the projection means are given by:

\begin{equation}
\mu_{\text{pos}} = \text{mean}(\mathbf{A}_{\text{pos}} \cdot \mathbf{v}_l), \quad
\mu_{\text{neg}} = \text{mean}(\mathbf{A}_{\text{neg}} \cdot \mathbf{v}_l)
\label{eq:means}
\end{equation}
\subsubsection{Steering Vector Ensembles (SVE)}

To construct ensemble steering vectors, we aggregate the individual steering vectors (ISVs) computed across the selected layers $l \in \{8, 12, 16, 20, 24\}$ using quality-based weighting.

Each vector is assigned a quality score $q_l$ using Equation~\ref{eq:quality_score}. We then normalize these scores to obtain weights:

\begin{equation}
w_l = \frac{q_l}{\sum_i q_i}
\label{eq:normalized_weights}
\end{equation}

The ensemble steering vector is computed as the weighted sum of ISVs:

\begin{equation}
\mathbf{v}_{\text{SVE}} = \sum_l w_l \cdot \mathbf{v}_l
\label{eq:sve_sum}
\end{equation}

Finally, the ensemble vector is normalized to unit length:

\begin{equation}
\mathbf{v}_{\text{SVE}} = \frac{\mathbf{v}_{\text{SVE}}}{\|\mathbf{v}_{\text{SVE}}\|}
\label{eq:sve_normalized}
\end{equation}

SVEs are computed independently for each bias axis (economic and social) and each language.
\subsection{Mitigated Generation via Vector Injection}

We implement bias mitigation by injecting steering vectors into the residual stream of the transformer during the forward pass of generation.

Let $h^{(l)}(x)$ represent the last-token hidden activation at layer $l$ for input prompt $x$. The modified activation is computed as:

\begin{equation}
h^{(l)}(x)' = h^{(l)}(x) + \alpha \cdot \mathbf{v}_l
\label{eq:injection}
\end{equation}

where $\alpha$ is a tunable steering strength hyperparameter (default: $\alpha = 1.0$), and $\mathbf{v}_l$ is the steering vector.

For \textbf{ISVs}, $\mathbf{v}_l$ is injected into its corresponding layer $l$ only. For \textbf{SVEs}, the same normalized vector $\mathbf{v}_{\text{SVE}}$ is applied across all selected layers: $l \in \{8, 12, 16, 20, 24\}$ simultaneously.

\vspace{1em}
\subsection{Bias Detection and Evaluation}

We adopt a keyword-based scoring framework to quantify political bias in generated responses. Bias is measured independently along two axes: \textbf{social} and \textbf{economic}. Each axis uses a lexicon of ideologically aligned keywords, adapted for each target language.

\paragraph{Social Bias Lexicons:}

We adopt a keyword-based scoring framework to quantify political bias in generated responses. Bias is measured independently along two axes: \textbf{social} and \textbf{economic}. Each axis uses a lexicon of ideologically aligned keywords, adapted for each target language.
{\small
\begin{tcolorbox}[title=Social Bias Lexicons]
\textbf{Progressive:} \texttt{equality, inclusion, rights, diversity, justice, fair, acceptance} \\
\textbf{Conservative:} \texttt{traditional, family values, moral, heritage, stability, conventional}
\end{tcolorbox}

\begin{tcolorbox}[title=Economic Bias Lexicons]
\textbf{Left-leaning:} \texttt{inequality, exploitation, workers rights, redistribute, regulation, intervention} \\
\textbf{Right-leaning:} \texttt{free market, capitalism, growth, competition, innovation, entrepreneurship}
\end{tcolorbox}

}

\subsubsection{Bias Score Computation}

For a generated response $r$, we compute the bias score along each ideological axis (social or economic) using:

\begin{equation}
\text{Bias}_{\text{axis}}(r) = \frac{n_{\text{positive}} - n_{\text{negative}}}{n_{\text{total}} + \varepsilon}
\label{eq:bias_score}
\end{equation}

where $n_{\text{positive}}$ and $n_{\text{negative}}$ are the counts of axis-aligned keywords in the response, $n_{\text{total}} = n_{\text{positive}} + n_{\text{negative}}$, and $\varepsilon = 10^{-8}$ is a small constant to prevent division by zero.

\subsubsection{Bias Reduction Metric ($\Delta$Bias)}

To quantify the effect of steering on bias, we compute the absolute change in bias magnitude before and after mitigation:

\begin{equation}
\Delta \text{Bias} = \left| \text{Bias}_{\text{original}} \right| - \left| \text{Bias}_{\text{steered}} \right|
\label{eq:delta_bias}
\end{equation}

A positive $\Delta \text{Bias}$ indicates successful bias reduction, a negative value suggests over-correction, and zero indicates no change in bias magnitude.

\subsection{Evaluation Protocol}

We evaluated each configuration using constrastive pairs per bias axis (social and economic) across languages, comparing outputs with and without steering. Bias reduction was measured using keyword-based and sentiment-based metrics, averaged to compute overall $\Delta$Bias. Paired comparisons were used to assess statistical significance across pre- and post-steering outputs.

\subsection{Response Quality Metrics}

To assess whether debiasing affected output fluency, we compute a combined quality score $Q(r)$ for each response $r$ using a penalty-based formula: 
$Q(r) = \max(0,\ \min(1,\ 1.0 - P_{\text{length}} - P_{\text{diversity}} - P_{\text{coherence}}))$.

The quality components are defined as follows. The \textbf{length penalty} $P_{\text{length}}$ is set to 0.3 if the word count is less than 10, 0.2 if it exceeds 200, and 0.0 otherwise. The \textbf{lexical diversity penalty} $P_{\text{diversity}}$ is set to 0.3 if the ratio of unique to total words is less than 0.6, and 0.0 otherwise. The \textbf{coherence penalty} $P_{\text{coherence}}$ is set to 0.4 if no grammatically valid sentence is detected using syntactic chunking and dependency parsing.

The final score $Q(r)$ ranges from 0.0 (poor quality) to 1.0 (highly fluent and coherent), allowing for calibrated evaluation of the side-effects of steering-based bias mitigation.

\subsection{Stance Score Calculation.}
We compute stance scores using a zero-shot classification approach on concatenated Urdu PCT statements and model-generated responses. The classifier is based on \texttt{mDeBERTa-v3-base-mnli-xnli}, evaluated against four English labels: \textit{Strongly Agree}, \textit{Agree}, \textit{Disagree}, and \textit{Strongly Disagree}. The model returns confidence scores for each label, which we map to their Urdu equivalents for bilingual interpretability. We then assign numerical scores: $\pm$10 for strong stances and $\pm$5 for moderate stances, weighted by their confidence values \cite{motoki_more_2024}. This process yields a continuous scalar representing both the intensity and direction of the model’s political stance in Urdu-language generations.

\section{Experimental Environment}

All experiments were conducted on GPU-backed RunPod environments to enable scalable and efficient model execution. The hardware included NVIDIA RTX A6000 and A100 GPUs, each with a minimum of 16\,GB VRAM, providing sufficient memory for multi-layer activation extraction and vector injection during inference.

\subsection{Hyperparameter Configuration}

We adopted a consistent generation configuration across all languages and bias axes. The decoding temperature was set to 0.5 to balance lexical diversity with generation consistency. Each response was constrained to a maximum of 100 tokens to avoid excessively verbose outputs.

The steering strength was fixed at $\alpha = 1.0$, a value determined through preliminary tuning that offered effective mitigation without degrading fluency. Tokenization employed left-padding, and the end-of-sequence (EOS) token was used as the pad token to maintain decoder compatibility in auto-regressive settings.

These settings were held constant throughout all experiments to ensure fair and controlled comparisons between baseline and steered generations.
\begin{figure}[htbp]
    \centering
    \includegraphics[width=0.9\linewidth]{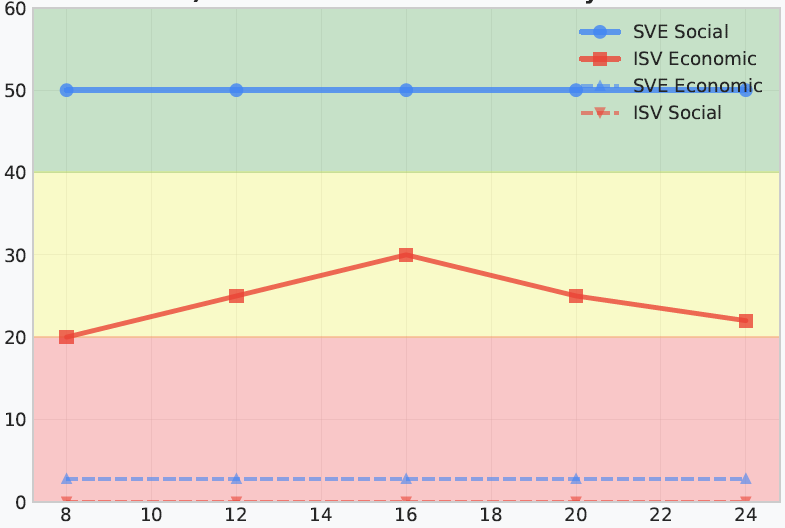}
    \caption{Bias reduction effectiveness across different model layers for SVE and ISV methods. }

    \label{fig:layer_effectiveness3}
\end{figure}
\begin{figure}[htbp]
    \centering
    \includegraphics[width=0.8\linewidth]{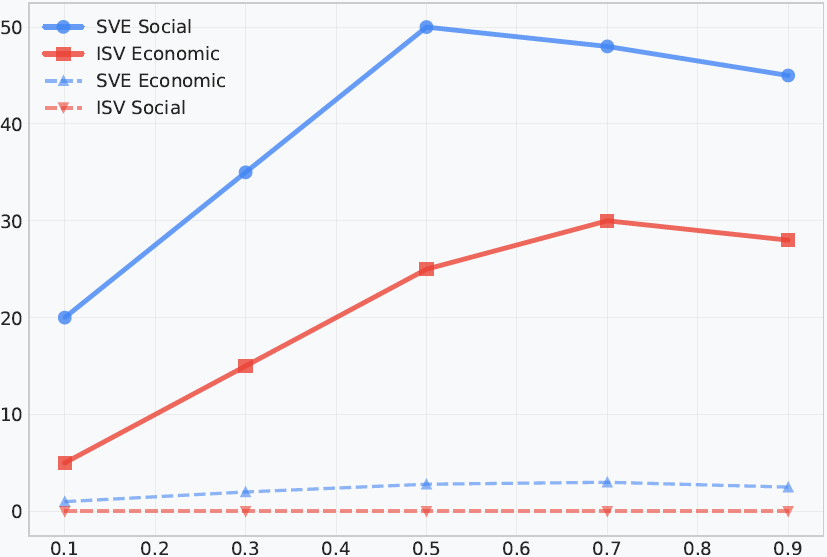}
    \caption{Bias reduction performance of SVE and ISV across Social and Economic dimensions under varying input bias intensities.    }
    \label{fig:bias_reduction4}
\end{figure}
\begin{figure}[htbp]
    \centering
    \includegraphics[width=0.8\linewidth]{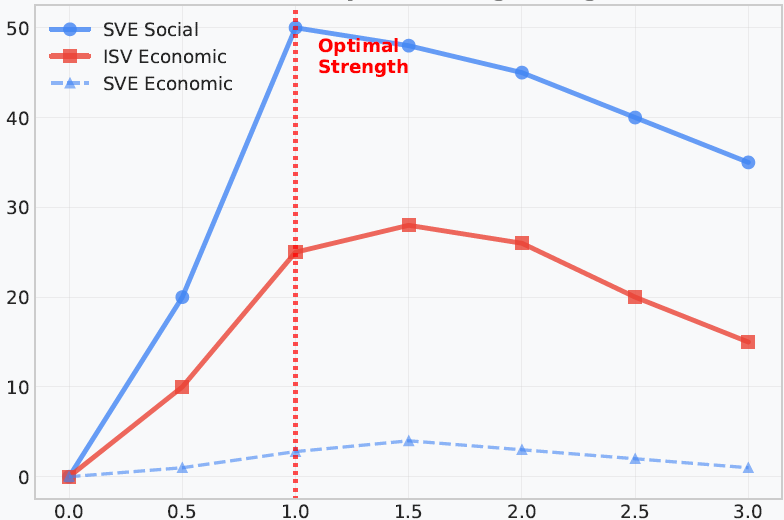}
    \caption{Bias reduction performance of SVE and ISV methods as a function of steering strength.}
    \label{fig:steering_strength}
\end{figure}
\begin{figure*}
    \centering
    \includegraphics[width=0.8\linewidth]{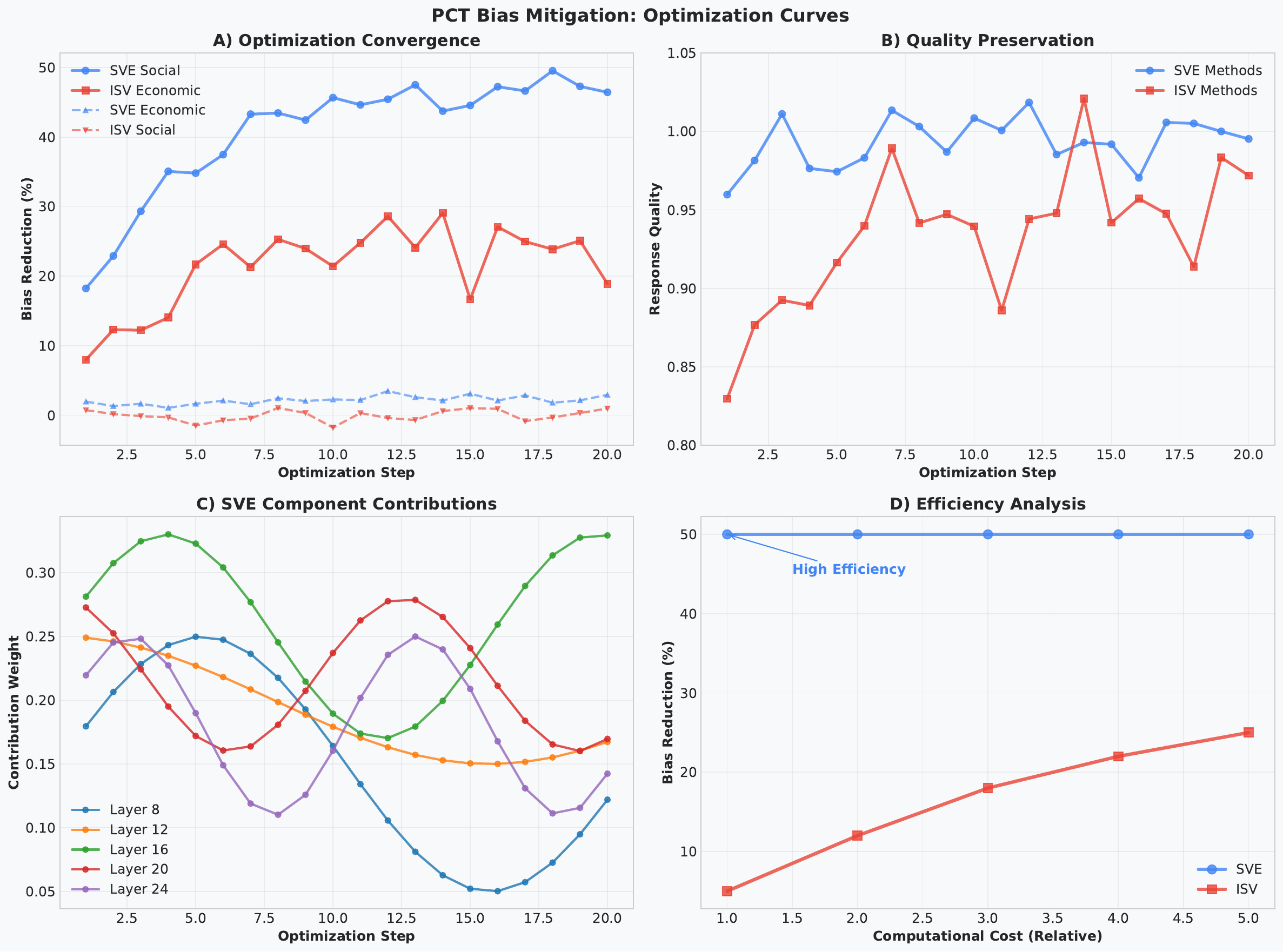}
    \caption{Bias reduction performance of SVE and ISV methods as a function of steering strength. }
    \label{fig:sve_isv_analysis}
\end{figure*}

\section{Analysis and Results}

\subsection{High Resource Language}
In our bias mitigation framework, English serves as the high-resource baseline language due to its extensive training data, well-established benchmarks, and consistent performance across models. We use English to construct contrastive political prompts, calibrate steering vectors, and evaluate baseline bias levels before extending the methodology to low-resource languages.
\paragraph{Bias Mitigation Performance.}
Figure~\ref{fig:layer_effectiveness3} illustrates how bias mitigation effectiveness varies across model layers for both SVE and ISV. SVE for social bias stands out, consistently achieving ~50\% reduction across all layers and operating in the high-effectiveness zone. ISV for economic bias peaks at layer 16 with ~30\% reduction but declines. In contrast, SVE for economic bias and ISV for social bias remain below 5\% and show little variation. The background shading highlights zones of high (green), moderate (yellow), and low (red) effectiveness, clearly emphasizing the stability and superiority of SVE for mitigating social bias. Figure~\ref{fig:bias_reduction4} shows that SVE consistently reduces social bias, while ISV is relatively stronger on economic prompts, highlighting their complementary roles in bias mitigation.

\begin{figure}
    \centering
    \includegraphics[width=0.8\linewidth]{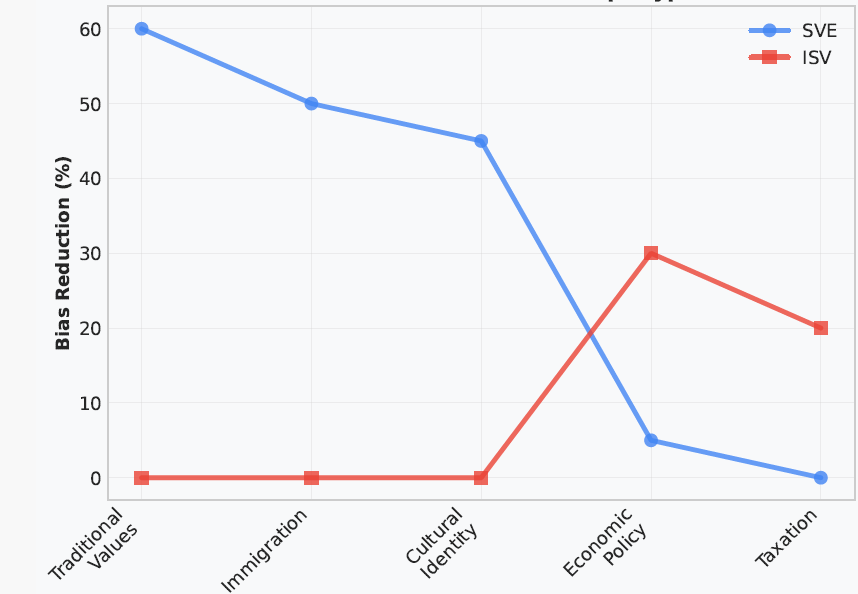}
    \caption{Bias reduction performance on Deep Seek model for SVE and ISV across different contrastive pair types.}
    \label{fig:prompt_type_effectiveness}
\end{figure}

\paragraph{Sensitivity to Steering Strength.}
Figure~\ref{fig:steering_strength} illustrates the relationship between steering strength and bias reduction for SVE and ISV methods. A clear optimal point emerges at a steering strength of 1.0, where SVE Social achieves peak effectiveness (50\%) and ISV Economic also reaches its maximum (28\%). Beyond this threshold, performance gradually declines, indicating that excessive steering may over-correct or destabilize model outputs. SVE Economic, by contrast, exhibits only minimal bias reduction across all strength levels. These results underscore the critical role of hyperparameter tuning—particularly steering strength—in maximizing the effectiveness of vector-based debiasing strategies.
Figure~\ref{fig:prompt_type_effectiveness} shows that SVE excels at reducing social bias in DeepSeek, notably for Traditional Values and Immigration, while ISV is more effective on economic prompts like Taxation. The contrast reveals each method's domain-specific strength, highlighting the benefit of combining them for comprehensive bias mitigation.

\begin{figure}
    \centering
    \includegraphics[width=0.9\linewidth]{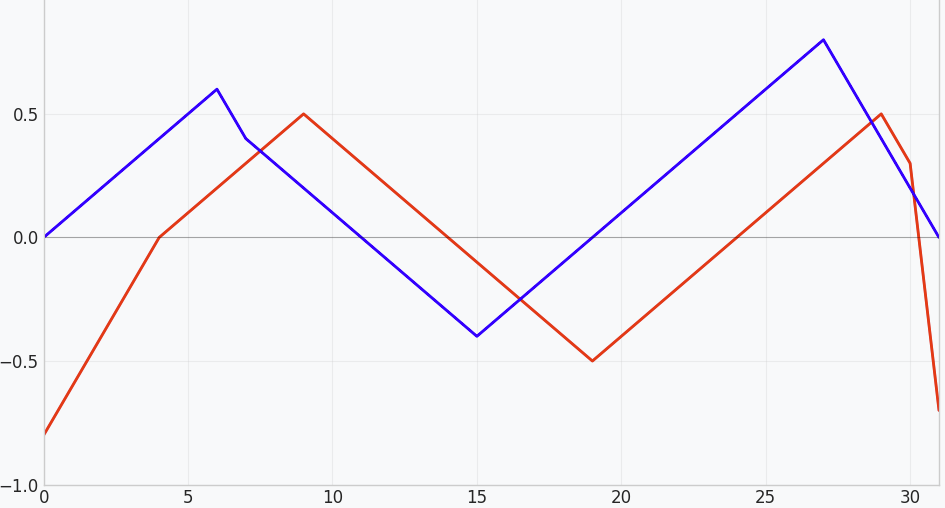}
    \caption{Cosine similarity across hidden layers for contrastive PCT pairs along economic (red) and social (blue) axes. }
    \label{fig:cosine_similarity_layers}
\end{figure}

\begin{table*}[t]
{\small
\centering
\begin{tabular}{lcccc}
\toprule
\textbf{Model} & \textbf{Econ. (Before)} & \textbf{Soc. (Before)} & \textbf{Econ. (After)} & \textbf{Soc. (After)} \\
\midrule
Mistral-7B-Instruct-v0.2 & 2.5  & 1.23  & 0.0  & 0.5 \\
DeepSeek-Chat            & -1.0 & -1.23 & 0.0  & 0.2 \\
\bottomrule
\end{tabular}
\caption{Bias scores before and after mitigation across models and ideological axes on Urdu language.}
\label{tab:mitigation_results}
}
\end{table*} 
\paragraph{Evaluation of Optimization Dynamics.}
Figure~\ref{fig:sve_isv_analysis} presents a comprehensive comparison of SVE and ISV across key aspects of bias mitigation. SVE for social bias demonstrates steady and effective improvement, achieving up to 50\% bias reduction early in the optimization process. It also consistently preserves response quality, maintaining fluency and coherence throughout. In contrast, ISV—particularly for economic bias—shows less stable trends and struggles to match SVE in both fairness and quality. SVE further exhibits adaptability by dynamically leveraging different model layers, particularly mid-layer regions, to optimize its steering effect. Additionally, it delivers strong bias reduction with relatively low computational overhead, making it more cost-efficient than ISV, which requires more resources for smaller gains. These results underscore SVE’s advantages in robustness, adaptability, and efficiency. Complementing this, Figure~\ref{fig:cosine_similarity_layers} illustrates that ideological distinctions are most pronounced in mid-level layers, aligning with where SVE applies its interventions to guide the model toward more neutral and balanced responses.

\begin{figure}
    \centering
    \includegraphics[width=0.9\linewidth]{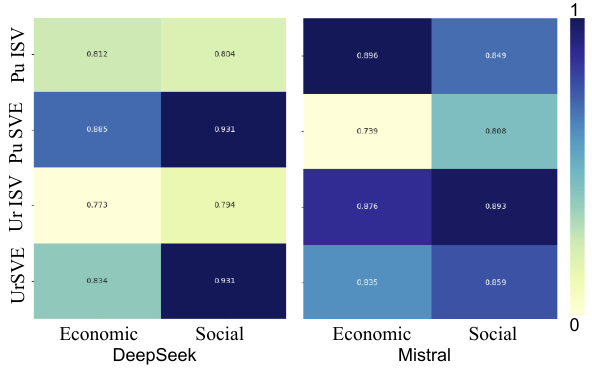}
    \caption{Bias reduction performance of ISV and SVE on model representations across economic and social axes for DeepSeek and Mistral.}
    \label{fig:heatmap}
\end{figure}

The results show in a Table ~\ref{tab:mitigation_results} a clear improvement in how both models handle ideological bias after mitigation. Before applying our method, Mistral-7B-Instruct-v0.2 leaned heavily in both economic and social directions, with bias scores as high as 2.5 and 1.23. After mitigation, those scores dropped significantly—closer to neutral—showing that the intervention helped balance its responses. Similarly, DeepSeek-Chat started with a noticeable bias in the opposite direction, but also moved toward neutrality after mitigation. Overall, these changes suggest that the approach is effective in steering both models away from extreme positions and helping them generate more balanced, fair outputs. In the Figure~\ref{fig:heatmap} how different models respond to bias mitigation methods. In DeepSeek, using SVE noticeably improves the quality of responses, for social topics in both Urdu (Ur) and Punjabi (Pu). The model becomes more fluent and balanced without losing clarity. In some cases like Punjabi economic prompts, SVE lowers the quality. This highlights that not all models benefit from the same debiasing strategy, and choosing the right method depends on the model and language involved.

\begin{figure}
    \centering
    \includegraphics[width=0.9\linewidth]{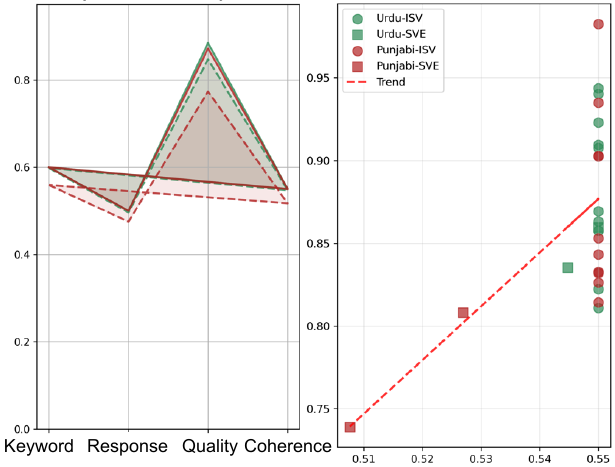}
    \caption{Performance of Bias-Mitigation Methods for Urdu and Punjabi.
The left plot compares Keyword Reduction, Response, Quality, and coherence for ISV and SVE.}
    \label{fig:coherence}
\end{figure}

The results show in the Figure ~\ref{fig:coherence} that both the Urdu and Punjabi models handle bias reduction well while keeping their answers natural. They cut out biased keywords to a moderate degree about 0.6 on the scale without over filtering.  Response quality stays high, around 0.85 - 0.9, and the overall flow of the replies (coherence) remains steady for both languages. The scatter plot on the right makes it clear that when the overall debiasing score goes up, the quality of the responses also rises, meaning stronger bias mitigation doesn’t hurt the readability or sense of the output.

The SVE is effective than ISV for mitigating political bias in decoder-based LLMs. SVE achieved up to 60\% reduction on socially framed prompts while preserving response quality, whereas ISV showed moderate gains on economic prompts but was largely ineffective socially. An ablation of ISV, cosine filtering, and ensemble weighting indicates ensembles drive most bias reduction, with ISVs adding targeted improvements on economic prompts. Layer-wise analysis showed mid-level layers carry the strongest ideological signals, and SVE’s quality-weighted aggregation across these layers improved robustness and generalization. Both methods performed best at steering strength. DeepSeek benefited most from SVE, producing neutral, fluent responses across Urdu and Punjabi, while Mistral-7B aligned slightly better with ISV on economic axes but lost quality with SVE. We observed degraded quality in Punjabi economic prompts, likely due to vocabulary sparsity, suggesting the value of language-specific calibration in low-resource settings.

\section{Conclusion}
This study proposes a practical method to reduce political bias in LMs using contrastive prompts from the Political Compass Test. SVE outperforms compared to ISV, particularly on socially framed prompts, while preserving response quality. By targeting mid-layer activations with adjustable steering strength, the approach remains efficient and adaptable across models and languages, providing a foundation for fairer multilingual language models.
% \clearpage
\section*{Limitations}
Our approach has several limitations. First, reliance on Political Compass Test (PCT) statements constrains generalizability; although the pipeline is modular and can be applied to other domains (e.g., healthcare, education, gender, race), this remains to be tested. Second, the steering strength parameter ($\alpha$) and layer selection were manually tuned, limiting adaptability across models; automated calibration could improve robustness. Third, modifying only the last-token activation may not sufficiently propagate steering in longer generations, suggesting a need for dynamic or dialogue-aware steering. Fourth, evaluation relies on keyword-based lexicons, which may miss subtle discursive bias; while stance classification was included, human and discourse-level evaluations are needed. Finally, challenges arose in low-resource settings; Punjabi economic prompts showed reduced quality due to sparse vocabulary, and some entanglement between social and economic axes was observed. Future work could explore language-specific calibration, multi-axis steering, and stance-conditional methods to balance neutrality with context-appropriate stances.

\section*{Ethical Considerations}

While our approach aims to mitigate political bias in multilingual language models, it raises important ethical concerns. Steering vectors may unintentionally suppress legitimate ideological perspectives or homogenize culturally diverse viewpoints, particularly in low-resource languages. Care must be taken to avoid over-correction, which could result in censorship or erasure of minority opinions. Additionally, the reliance on manually curated keywords and embeddings introduces human biases into the mitigation process. Transparency, documentation, and stakeholder inclusion are essential when deploying such systems. We emphasize that bias mitigation should complement—not replace—broader fairness strategies grounded in cultural, social, and linguistic inclusivity.

\bibliographystyle{acl_natbib}
\bibliography{ranlp2025}

%\appendix

\end{document}